\documentclass{svproc}
\usepackage{url,graphicx, dsfont}

\usepackage{amsmath,amssymb,amsfonts,xspace,epsfig,stackrel,latexsym,wrapfig,verbatim,psfrag,rotating}
\usepackage[top=1in, bottom=1in, left=1in, right=1in]{geometry}
\RequirePackage[colorlinks,citecolor=blue,urlcolor=blue]{hyperref}

\usepackage{subfigure}

\newcommand{\twopiece}[5][0]{
    \ifcase#1
        \left\{\begin{array}{ll}{#2}&{\text{ if } #3}\\{#4}&{\text{ if } #5}\end{array}\right.
    \else
        \left\{\begin{array}{ll}{#2}&{\text{ if } #3}\vspace{#1pt}\\{#4}&{\text{ if } #5}\end{array}\right.
\fi}

\newcommand{\threepiece}[7][0]{ 
    \ifcase#1
        \left\{\begin{array}{ll}{#2}&{\text{ if } #3}\\{#4}&{\text{ if } #5}\\{#6}&{\text{ if } #7}\end{array}\right.
    \else
        \left\{\begin{array}{ll}{#2}&{\text{ if } #3}\vspace{#1pt}\\{#4}&{\text{ if } #5}\vspace{#1pt}\\{#6}&{\text{ if } #7}\end{array}\right.
\fi}

\begin{document}

\mainmatter              
\title{Ae$^2$I: A Double Autoencoder for Imputation of Missing Values}
\titlerunning{Double Autoencoder}  
%
\author{Fuchang Gao}
\authorrunning{Fuchang Gao} 
%
\tocauthor{Fuchang Gao}
\institute{Department Mathematics and Statistical Science, University of Idaho\\ 875 Perimeter Drive, MS 1103, Moscow, ID 83844-1103, USA,\\
\email{fuchang@uidaho.edu},\\ WWW home page:
\texttt{https://www.webpages.uidaho.edu/~fuchang/}
}

\maketitle              

\begin{abstract}
The most common strategy of imputing missing values in a table is to study either the column-column relationship or the row-row relationship of the data table, then use the relationship to impute the missing values based on the non-missing values from other columns of the same row, or from the other rows of the same column.
This paper introduces a double autoencoder for imputation (Ae$^2$I) that simultaneously and collaboratively uses both row-row relationship and column-column relationship to impute the missing values. Empirical tests on Movielens 1M dataset demonstrated that Ae$^2$I outperforms the current state-of-the-art models for recommender systems by a significant margin.

\end{abstract}
\section{Introduction}
From medical research to recommender systems, missing data are ubiquitous. In many cases, the missing values need to be imputed before the dataset can be used. Indeed, imputation of missing values has been studied for decades \cite{Rubin}, and many sophisticated methods of imputation have been developed using statistical or machine learning approaches \cite{schafer1999multiple},\cite{imputation_survey}. One basic strategy is to impute missing values with a weighted average of some selected non-missing values. The simplest is Mean Imputation, which imputes a missing value by the simple average of all non-missing values from the same row or same column.
This simple imputation decreases the variance and alters the covariance structure of the original data. An improvement is to use the average of selective entries only. Along this line, the hot deck methods use some specific assumptions on missing patterns to determine a set of auxiliary variables based on which to match donors for each missing entry \cite{hotdeck}. The $k$-nearest neighbor (KNN) imputation \cite{knn} method uses a metric to define the similarity between two rows or columns. For each missing entry, the $k$ most similar rows or columns with non-missing values in the corresponding entry are selected. Because there are many different ways to introduce the similarity metric,  KNN enables more sophisticated ways than hot deck. The selection of the similarity metric is of key importance to the performance of the model. Various modifications and refinements have led to many variants of KNN, such as Sequential KNN \cite{seq-knn} and Iterative KNN \cite{iter-knn}.

The basic assumption behind these selection methods is that each missing value can be viewed as a function of some non-missing values in the same row or same column. One specific example is to assume that column (or row) variables satisfy some multivariate relationship. For instance, the Maximum Likelihood methods \cite{Anderson} assume a joint probability distribution of these variables, say a multi-normal distribution, then use non-missing values to find the best parameters for that joint distribution. Regression methods \cite{missing-data} assume a conditional probability distribution of one variable given a sequence of variables with complete data, then conduct univariate imputations sequentially until convergence. To avoid noise introduced by imputation, Multiple Imputation (\cite{Rubin1987} \cite{MI}) techniques were introduced. A popular example is Multiple Imputation by Chained Equations (MICE) (c.f. \cite{chained-eq}).

To assume a specific joint distribution of all the variables is very strict. A generalization is to assume that these variables can be expressed as a function of some smaller set of latent variables. The latter assumption can be justified by low-rank approximation. Indeed, if we view the data table as a rectangular matrix $X$, and factorize the matrix using singular value decomposition into $X=U\Sigma V^T$, where $\Sigma$ is a diagonal matrix with singular values of $X$ in its diagonal. By replacing the small singular values in $\Sigma$ with 0 to create a new diagonal matrix $\widehat{\Sigma}$ with $k$ non-zero diagonal entries, we can approximate $X$ by a low-rank matrix $\widehat{X}:=U\widehat{\Sigma} V^T$. If we denote by $H$ the first $k$-column of the matrix $U\sqrt{\Sigma}$, and $W$ the first $k$ rows of the matrix $\sqrt{\Sigma} V^T$, we can approximate $X\approx HW$. If we view the $k$ row vectors of $W$ as latent variables, then each column variables of $X$ can be approximately expressed as a linear combination of these $k$ latent variables.  Matrix Factorization (MF) (e.g. \cite{MF1} \cite{MF2} \cite{MF3}) methods have been developed from this idea.

Note that the latent variables in MF are linear functions of the original variables. Extending from linear to non-linear functions is the main contribution of Collaborative Filtering (CF) (e.g. \cite{J-NCF}). Closely related to CF are autoencoders. Unlike CF that starts with random low-rank matrices $H$ and $W$ to recover $X$, autoencoders start with $X$ with missing values filled with 0, and generate low-rank matrices, and then use them to recover $X$. Starting from $X$ takes more computations than directly starting with two random low-rank matrices, but being paid off with a more targeted generation of low-rank matrices. Because the idea of autoencoder is directly related to the current work, let us take a closer look.

An autoencoder is a fully-connected neural network consisting of two parts. The first part is called an encoder, which transforms $d$-dimensional input vectors to $k\ll d$ dimensional vectors through one or several layers. The second part is called a decoder, which transforms the $k$-dimensional vectors back to $d$-dimensional vectors. If the input vector $x$ is a column vector, then the encoder transformation can be expressed as
$$\tilde{x} = W_r\sigma \left(W_{r-1}\left(\cdots \sigma\left(W_1x+b_1\right)\cdots\right)+b_{r-1}\right)+b_r,$$
where $W_i$, $b_i$ are the weights and bias in the $i$-th layer. The encoder transforms the input $d$ variables into $k$ latent variables.
The decoder transformation can be expressed as
$$\hat{x} = W_s\sigma \left(W_{s-1}\left(\cdots \sigma\left(W_{r+1}\tilde{x}+b_{r+1}\right)\cdots\right)+b_{s-1}\right)+b_s,$$
which decodes the $k$ latent variables back to the original $d$ variables.
Thus, an autoencoder can be viewed as non-linear matrix factorization.

Many variants of autoencoder models have been developed. The most popular models include Autoencoder Recommender  (AutoRec) \cite{AutoRec}, Denoising Autoencoder (DAE) \cite{DAE}, Multiple Imputation Denoising Autoencoder \cite{MIDA} \cite{MIDA2}, Variational Autoencoder \cite{AE1} \cite{AE2}, and etc. For example,  DAEs have put up strong performances in imputing a wide range of datasets, from electronic health records to scRNA-seq data and traffic data. While AutoRec has great performance on many benchmark datasets for collaborative filtering, including Movielens-1M, Movielens-10M \cite{Movielens}.

In spite of their great performance, there is an obvious drawback in current autoencoder models for imputation, that is, they can only use column-column relationship or row-row relationship, not both. Observed the importance of using both row-row relationship and column-column relationships, Joint Neural Collaborative Filtering (J-NCF) was proposed in \cite{J-NCF}, in which column-column relationship was used to build a user-based autoencoder and row-row relationship was used to build an item-based autoencoder. The two autoencoders were trained separately, and the average of the two predictions was used for the final prediction. This idea was also used in Badsha {\it et al}. \cite{LATE} on single cell gene expression research. In their ``LATE combined" model, they alternately apply an autoencoder on row-row relationships, and an autoencoder on column-column relationships. The result was better than each individual autoencoder. However, because one relationship was significantly stronger than the other, the alternating training converged slowly, and the improvement was marginal.

There are also many other imputation methods such as the ones based on Generative Adversarial Networks (e.g. \cite{GAN1}, \cite{GAN2}, \cite{GAN3}). Because they are not directly related to the current work, we will not discuss them here.

In this paper, we introduce a double autoencoder with two branches, one of which uses row-row relationship and the other column-column relationship. Unlike J-NCF (\cite{J-NCF}) and ``LATE combined'' (\cite{LATE}, in which two autoencoders were trained separately, in our model, the two branches are connected in the last layers. By penalizing the discrepancy on the cross intersections of the two outputs, information gathered from one branch is passed to the other, enabling both branches to make better predictions.

\section{Methods}

To better introduce our methods, we represent the data as an $m\times n$ real-values matrix with missing valued entered as 0. The $m$ rows represents different observers, and the $n$ column represents different items (variables).
\subsection{Architecture}
Our model consists of two connected autoencoders (or branches), the row autoencoder and the column autoencoder. The row autoencoder is trained on row vectors of the matrix. By a nested sequence of column operations and non-linear transforms, the row autoencoder outputs an $m\times n$ matrix $X$ with the missing entries filled using the learnt column-column relationship. The column autoencoder is trained on column vectors of the matrix. By a nested sequence of row operations and non-linear transforms, the column autoencoder also outputs an $m\times n$ matrix with the missing entries filled in using the learnt row-row relationship. The two autoencoders are connected by the output layers. Figure 1 illustrates the architecture of the double autoencoder.
\begin{figure}[h!]
\centering
\includegraphics[width=4.5in]{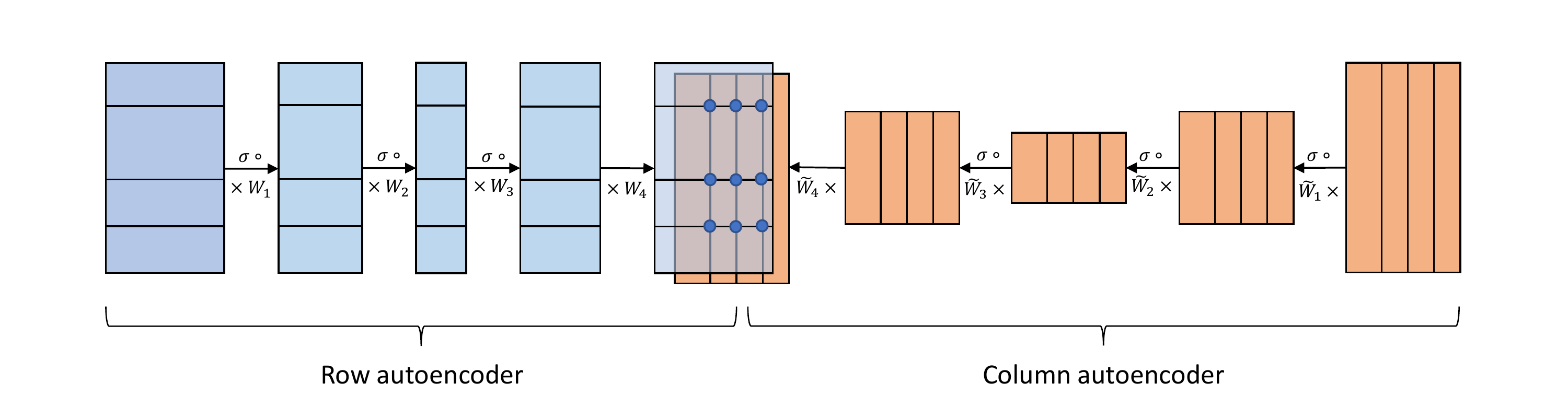}

\caption{The architecture of a double autoencoder.}
\end{figure}
The two antoencoders are trained simultaneously, and collaboratively. During each iteration, a random list $I$ of row vectors of $X$ are fed to the row autoencoders, and a random list $J$ of column vectors of $X$ are fed to the column autoencoder. Let $\hat{x}_{ij}$, $1\le j\le n$ be the output of the row autoencoder when $i$-th row of $X$ is fed into it, we define the loss function
$${\rm Loss_{row}}=\frac{\sum_{i\in I}\sum_{j=1}^n(\hat{x}_{ij}-x_{ij})^2*\mathds{1}(x_{ij} \text{ is not missing})}{\sum_{i\in I}\sum_{j=1}^n\mathds{1}(x_{ij} \text{ is not missing})}.$$
Similarly,
 let $\tilde{x}_{ij}$, $1\le  i\le m$ be the output of the column autoencoder when $j$-th column of $X$ is fed into it, we define the loss function
$${\rm Loss_{col}}=\frac{\sum_{j\in J}\sum_{i=1}^m(\hat{x}_{ij}-x_{ij})^2*\mathds{1}(x_{ij} \text{ is not missing})}{\sum_{j\in J}\sum_{i=1}^m\mathds{1}(x_{ij} \text{ is not missing})}.$$

\subsection{Connecting the Two Autoencoders}
The key component of the model is the added penalty function that ties the two autoencoders together. Because the two autoencoders are meant to impute the same matrix, at each iteration of training,  if $I$ row vectors are fed to the row autoencoder, and $J$ column vectors are fed in to column autoencoder, the values of the outputs of the two autoenconders are meant to be equal at the cross interactions of these $I$ rows and $J$ columns. This motivates us to introduce the cross loss as

$${\rm Loss_{cross}} = \frac{\sum_{i\in I}\sum_{j\in J} (\hat{x}_{ij}-\tilde{x}_{ij})^2}{|I|\cdot |J|},$$
where $|I|$ and $|J|$ mean the cardinality of $I$ and $J$ respectively.

\subsection{Consecutive Weight Regularization}
Weight regularization is important to reduce overfitting of neural networks. The most common weight regularization is probably the $L_2$-regularization. Here we use the consecutive weight regularization technique that has been recently developed in \cite{Gao-consecutive}. 
If we denote by $W_k(i,j)$ the $i$-th row $j$-th column entry of the weight matrix $W_k$ in $k$-th layer, then the consecutive weight regularization can be defined by
$${\rm Loss_{weight}}= \sum_{k=1}^s\sum_{h}\left|\sum_{i}W_{k+1}(i,h)^2-\sum_{j}W_k(h,j)^2\right|,$$
After taking this weight regularization into consideration, the total loss function of the double autoencoder is
\begin{equation}
L = {\rm Loss_{row}}+{\rm Loss_{col}}+\lambda {\rm Loss_{cross}}+\mu {\rm Loss_{weight}},\label{loss}
\end{equation}
where $\lambda$ is a penalty constant and $\mu$ is the consecutive weight regularization constant.

\subsection{Customizing Activation Functions}
It is known that almost all non-linear function can be used as an activation functions. The performances of activation functions, however, can be very different, and often depend the datasets and the model hyperparameters. There are several dozens of activation functions that are commonly used. In image classification, ReLU and Leaky ReLU are popular choices. For autoencoders for recommender systems, AutoRec \cite{AutoRec} reported that sigmoid function performed better than ReLU. In recurrent neural networks, hyperbolic tangent is more commonly used.

For almost all the existing activation functions, when $x\to\infty$, the functions are either bounded (such as in Sigmoid, Hyperbolic tangent, Binary Step), or grow linearly (such as in ReLU, Leaky ReLU, PReLU, ELU, GELU, SELU, SiLU, Softplus, Mish).  To take into consideration of the need for activation functions with different squashness at $x\to \infty$, and motivated by the even Seagull activation function constructed in \cite{Gao-Zhang}, we introduce a family of activation functions: $\sigma_\alpha(x)=\log(1+|x|^\alpha)$ for $\alpha>0$, $\sigma_\alpha^{\pm} ={\rm sign}(x)\log(1+|x|^\alpha)$ and $\sigma_\alpha^{+} ={\rm sign}(x)\log(1+(\max\{x,0\})^\alpha)$. Because these functions behavior like $\alpha\log x$ for large $x$, we call this family of activation functions as logarithmic growth family. These activation functions fill the gap between bounded activation functions and linear growth activation functions. The function $\log(1+x^2)$ was called Seagull in \cite{Gao-Zhang}, as its graph looks like a flying seagull. Here we call the function ${\rm sign}(x)\log(1+|x|)$ as linear logarithmic unit (LLU). Figure 2 illustrates the graph of these two functions.
\begin{figure}[h!]
\centering
  \includegraphics[width=4.5in]{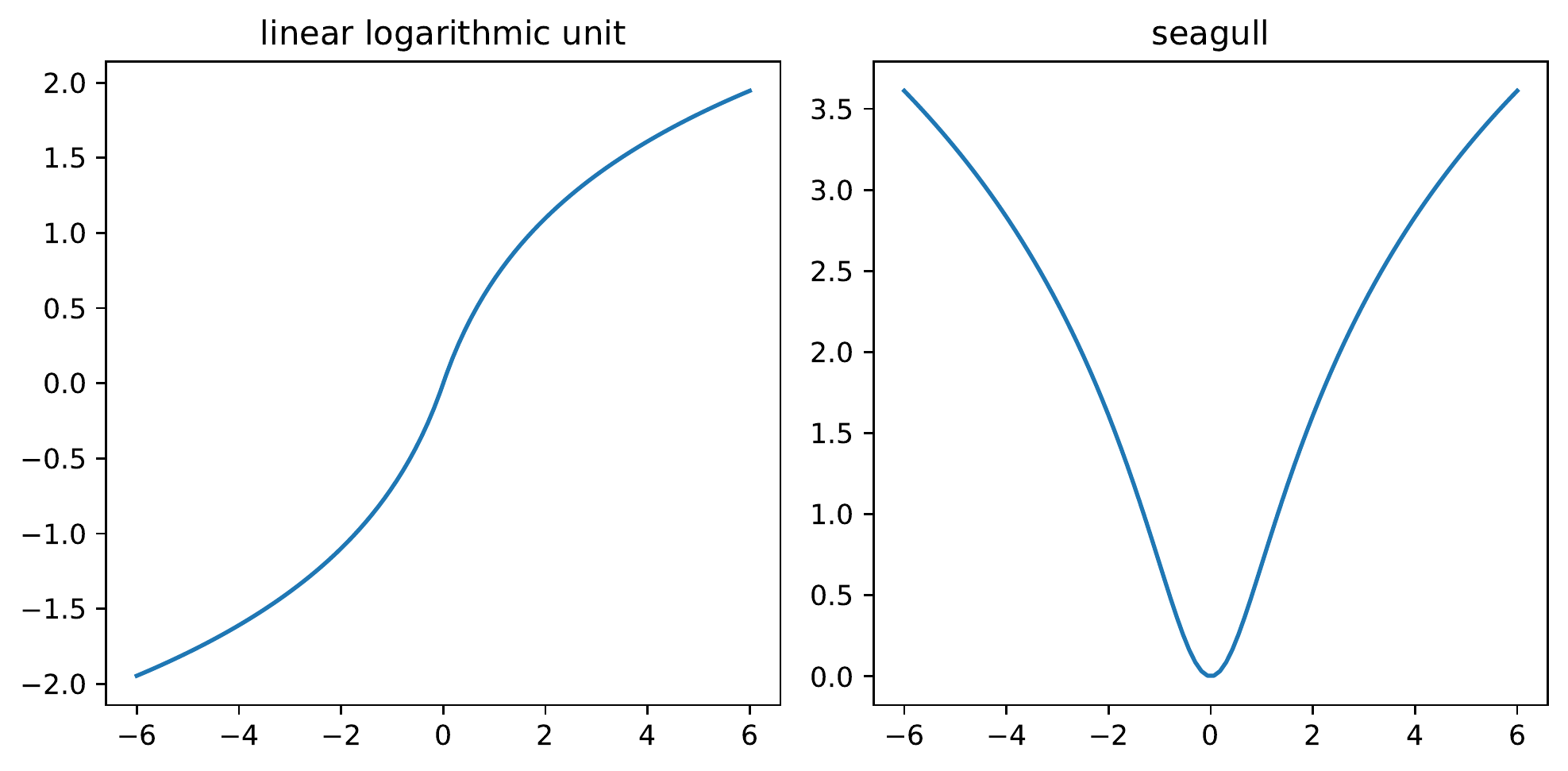}
  \caption{The graphs of activation functions with logarithmic growth}
\end{figure}

\section{Empirical Testing}
\subsection{Data Set}
To test the effectiveness of our model, we used the Movielens 100K  (ML-100K), and Movielens 1M (ML-1M) datasets \cite{Movielens}, which are popular stable benchmark datasets for recommendation systems. ML-100K contains about 100K anonymous ratings of 1682 movies by about 943 users. The data sparsity of 6.0\%. ML-1M contains about 1 million anonymous ratings of 3706 movies by about 6040 users. The sparsity of the matrix is 4.47\%. For ML-100K, we use the canonical u1.base for training, and u1.test for testing. For  ML-1M, we use random 90:10 split for train/test.
\subsection{Baseline Models}
We compared our model with 12 models which are either in the top 8 scorers for ML-1M, or in the top 6 scorer for ML-100K without using side information. These 12 models include: GLocal-K \cite{GLocal-K}, Sparse FC \cite{SparseFC}, CF-NADE \cite{CF-NADE}, I-AutoRec\cite{AutoRec}, GC-MC \cite{GC-MC}, I-CFN \cite{CFN}, BST \cite{BST}, NNMF \cite{NNMF}, GraphRec \cite{GraphRec}, IGMC \cite{IGMC}, Self-Supervised Exchangeable Model and Factorized EAE \cite{Self}. GLocal-K is a Global-Local Kernel-based matrix completion framework that generalizes and represents a high-dimensional sparse user-item matrix entry into a low dimensional space with a small number of important features. Sparse FC uses a neural network architecture in which weight matrices are re-parametrized in terms of low-dimensional vectors, interacting through kernel functions.  CF-NADE is a neural autoregressive architecture for CF tasks, which is inspired by the Restricted Boltzmann Machine (RBM) based CF model and the Neural Autoregressive Distribution Estimator (NADE). I-AutoRec is an autoencoder based model that uses item embeddings in the encoder. GC-MC is a graph auto-encoder framework based on differentiable message passing on the bipartite interaction graph. I-CFN is a hybrid model that enhances an autoencoder by using a loss function adapted to input data with missing values, and by incorporating side information. BST uses a transformer model to capture the sequential signals underlying users' behavior sequences followed by MLP. NNMF is a neural network matrix factorization model that replaces inner products in matrix factorization by a neural network. GraphRec uses a bipartite graph and constructs generic user and item attributes via the Laplacian of the user-item co-occurrence graph. IGMC is an inductive matrix completion model based on graph neural networks. Self-Supervised Exchangeable Model and Factorized EAE are deep learning models that use a parameter-sharing scheme.

\subsection{Implementation}
We experimented the model with the sizes of the hidden layers to be of the form $[m, k, m]$, for $m\in \{300,500,700\}$ and  $k\in \{50, 70, 200, 250\}$ in both autoencoders. For ML-1M, which contains more training data, a larger $k$ gave a slightly better result, and we chose $[500,250,500]$. For ML-100K which has relatively less training data, a smaller $k$ performed slightly better. We chose $[700, 70, 700]$.

We also experimented with 11 activation functions (Sigmoid, Tanh, ReLU, PReLU, ELU, GELU, SELU, SiLU, Softplus, Seagull, LLU). Sigmoid and Tanh are from the bounded family, ReLU and Softplus are from the linear growth family, and Seagull (i.e. $\log(1+x^2)$) and LLU (i.e. ${\rm sign}(x)\log(1+|x|)$) are from the logarithmic growth family. The functions from the same family had similar performance for both datasets. But activation functions from the different families reveal more noticeable difference. For ML-100K, the ones from linear growth family performed better, and we chose Softplus. For the ML-1M, the ones from the logarithmic growth family perform the best, and we chose Seagull.

ADAM optimizer was used to minimize the loss function (\ref{loss}) starting with learning rate 0.001.
The cross intersection penalty constant $\lambda$ was chosen as 0.001 for ML-100K and 0.004 for ML-1M. The consecutive-weight regularization constant $\mu$ was chosen as 0.001 for ML-100K and 0.002 for ML-1M. Batch sizes for the row autoencoder and the column autoencoder were chosen so that the entire dataset was divided into $60$ equal sized batches for ML-100K and $50$ equal sized batches for ML-1M.

To tune the hyperparameters, we used $k$-folder cross validation with $k=5,10,20,30$ folders. Bigger $k$ has a slightly better performance. However, after $k=30$, the gain diminished. Each folder was training 50 epochs with early stop with tolerance 5 determined by the weighted RMSE on the validation set, that was defined by
$$\hat{Y}_{\rm avg}=\frac{{\rm MSE_{col}}\hat{Y}_{\rm row}+{\rm MSE_{row}} \hat{Y}_{\rm col}}{{\rm MSE_{row}}+{\rm MSE_{col}}},$$
where ${\rm MES_{row}}$ and ${\rm MSE_{col}}$ are the mean square errors of the row and column autoencoders on the validation set. After an early stop, the learning rate, weight regularization strength, and cross-intersection discrepancy penalty are multiplied by a factor of $1/6, 1/3, 2$ respectively for ML-100K, and $1/2, 1/3,1/2$ respectively for ML-1M. The training was resumed from the best saved model. This process was repeated twice for ML-100K and 3 times for ML-1M. After the training of each folder completed, a prediction on the test set was made and saved. The average prediction of all $k$ folders was used as the final prediction.

\section{Results}
We compared our model with 12 baseline models that are either the top 8 scorers for ML-1M or the top 6 scorers for ML-100K without using side information. The results are in Table 1.
\begin{table}[h]
\centering
\caption{  Comparison of RMSE test results. The best results are highlighted in bold. All RMSE results of the baseline models are from the respective
papers cited in the first column.}

\begin{tabular}{lcc}
  \hline
  { Model} & {Movielens 100K} & { Movielens 1M} \\
  \hline
  GLocal-K \cite{GLocal-K}& 0.890 & 0.822 \\
  Sparse FC \cite{SparseFC}& - & 0.824 \\
  CF-NADE \cite{CF-NADE}& - & 0.829 \\
  I-AutoRec \cite{AutoRec}& - & 0.831 \\
  GC-MC \cite{GC-MC}& 0.910 & 0.832 \\
  I-CFN \cite{CFN}& - & 0.8321 \\
  BST \cite{BST}& - & 0.8401 \\
  NNMF \cite{NNMF}& -& 0.843 \\
  GraphRec \cite{GraphRec}& 0.901 & - \\
  IGMC \cite{IGMC}& 0.905& -\\
  Self-Supervised Exchangeable \cite{Self}& 0.910 & - \\
  Factorized EAE \cite{Self} & 0.920 & - \\
  Ae$^2$I (current) & {\bf 0.887} & {\bf 0.817}\\
   \hline
\end{tabular}
\end{table}

Our testing on the ML-100K (see Figure 3-left) and ML-1M datasets (data not shown) revealed that consecutive weight regularization helped to reduce the RMSE for the test error.
\begin{figure}[h!]
\centering
  \includegraphics[width=4.5in]{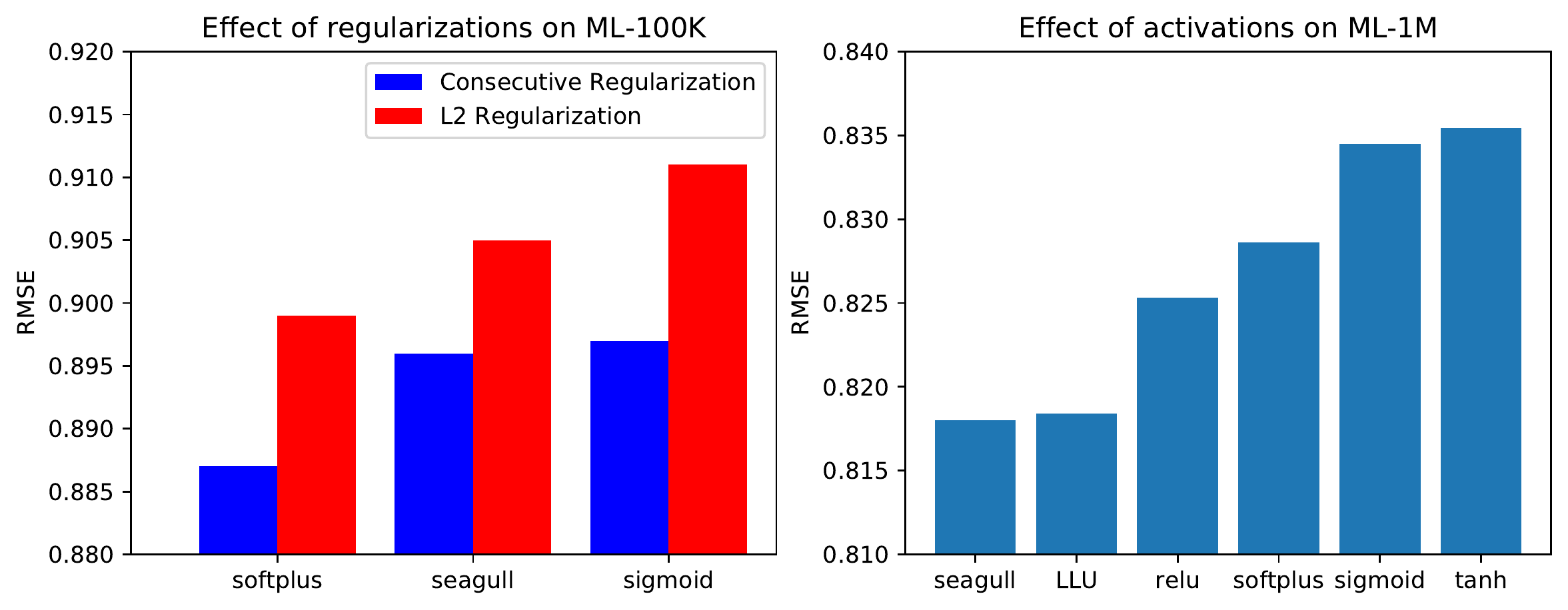}
  \caption{Effect of Regularization and Activation on RMSE}
\end{figure}

The effect of activations functions on the model performance depends on the dataset and model hyperparameters. For both ML-100K and ML-1M datasets, we observed that the activation functions with the same growth family tends to behave similarly, while activation functions from different growth families may have more noticeable difference. For ML-100K dataset, Softplus (from the linear growth family) performed the best. While for the ML-1M dataset, the logarithmic growth family (represented by Seagull and LLU) performed the best, see Figure 3-right. 

\section{Discussion}
In this paper, besides the double autoencoder, we used consecutive weight regularization, and demonstrated that it could be more effective than the usual $L_2$ regularization. We also noticed that under the new regularization, the training is more sensitive to learning rate and choice of activation function. Thus, an appropriate learning rate adjustment and customization of activation function appeared to be necessary. This may be due to the increased number of constraints on the weights posted by the consecutive weight regularization, which seems to have changed the distribution of the magnitudes of the weights, and thus, using a customized activation functions other than the commonly used linear growth family may take some advantages, as demonstrated by Figure 3-right.

\paragraph{Acknowledgement}
 We would like to acknowledge the support from the National Science Foundation grants OCA-1940270 and 2019609. 
%
%

\vskip 0.2in
\bibliographystyle{spmpsci}
\bibliography{Ae2I-reference}
\end{document}